\documentclass[10pt, a4paper]{article}

\usepackage[]{lrec-coling2024} 

\usepackage{tabularx}
\usepackage{booktabs}
\usepackage{adjustbox}
\usepackage{caption}
\usepackage{comment}

\title{Challenging Negative Gender Stereotypes: A Study on the Effectiveness of Automated Counter-Stereotypes}

\name{Isar Nejadgholi,$^1$, Kathleen C. Fraser,$^1$ Anna Kerkhof$^2$, Svetlana Kiritchenko$^1$ \\} 

\address{ $^1$National Research Council Canada, Ottawa, Canada \\$^2$ifo Institute for Economic Research and University of Munich, Munich, Germany \\ 
\small \texttt{\{isar.nejadgholi, kathleen.fraser, svetlana.kiritchenko\}@nrc-cnrc.gc.ca,}\\ \small \texttt{kerkhof@ifo.de}\\
}

\abstract{
\textit{\textbf{Content Warning:} This paper presents examples of gender stereotypes that may be offensive or upsetting.}\\
\newline
Gender stereotypes are pervasive beliefs about individuals based on their gender that play a significant role in shaping societal attitudes, behaviours, and even opportunities. Recognizing the negative implications of gender stereotypes, particularly in online communications, this study investigates eleven strategies to automatically counteract and challenge these views. 
We present AI-generated gender-based counter-stereotypes to (self-identified) male and female study participants and ask them to assess their offensiveness, plausibility, and potential effectiveness. 
The strategies of counter-facts and broadening universals (i.e., stating that anyone can have a trait regardless of group membership) emerged as the most robust approaches, while humour, perspective-taking, counter-examples, and empathy for the speaker were perceived as less effective. 
Also, the differences in ratings were more pronounced for stereotypes about the different targets than between the genders of the raters. 
Alarmingly, many AI-generated counter-stereotypes were perceived as offensive and/or implausible.
Our analysis and the collected dataset offer foundational insight into counter-stereotype generation, guiding future efforts to develop strategies that effectively challenge gender stereotypes in online interactions. 
 \\ \newline }

\begin{document}

\maketitleabstract

\section{Introduction}

Stereotypes involve attributing certain characteristics to a person purely on the basis of their perceived membership in a certain social category, often defined by demographic features such as race, ethnicity, age, or religious affiliation. In particular, perceived \textit{gender}\footnote{This study is limited to binary gender stereotypes; however, we acknowledge the prevalence of harmful stereotypes about non-binary individuals and believe the complexity of the issue warrants a dedicated study.} continues to be one of the most salient features by which these conscious and subconscious social categorizations are made, despite growing recognition that gender is not necessarily apparent from a person’s appearance, is not a binary categorization, and in most cases is not relevant to the situation \cite{ellemers2018gender}. Children as young as five years-old show a strong tendency to sort people into male and female categories \cite{aina2011does}, and as young as six years-old make assumptions about a person’s intelligence \cite{bian2017gender} based on this categorization. 
Gender stereotypes can be harmful to people of all genders, as they define particular expectations for how people can and should behave, regardless of individual strengths and weaknesses. Thus, from a young age, girls and women are expected to be friendly, nurturing, deferential, and concerned with presenting a feminine appearance, while boys and men are expected to be strong, competitive, and unemotional.

Studies show that gender stereotypes lead to biased perceptions of women's intellectual and leadership abilities, limiting their career opportunities; for example, when the same CV is submitted with a typically male versus female name, the male candidates are judged as more competent \cite{moss2012science}. Similar examples of gender discrimination against women in the professional sphere are unfortunately quite common \cite{correll2007getting, buffington2016stem, bobbitt2011gender}. 
Meanwhile, the pressure experienced by men to conform to masculine stereotypes can lead to impaired mental health and substance abuse \cite{wong2017meta}. Stereotypical beliefs about masculine gender roles also lead to a lack of help-seeking by male victims of intimate partner violence and sexual assault \cite{bates2019impact}.

Stereotypes are reinforced by repeated exposure. On the other hand, stereotypical associations can be weakened by exposure to \textit{counter-stereotypes} or information that disrupts or challenges the stereotype. Several counter-strategies can be employed, such as providing factual information to contradict the stereotype, asking questions to motivate critical thinking, or encouraging the speaker to ``put themselves in the target group's shoes''.  While many of the social psychology studies on counter-stereotypes involve in-person interventions, we are interested in the question of how to effectively generate counter-stereotypes in online spaces, such as on social media platforms, where such content is prevalent \cite{felmlee2020sexist, kerkhof2023gender}. To that end, we present the results of an online study to assess whether generative AI technology (in our case, ChatGPT) can be used to generate appropriate and plausible counter-stereotypes and which counter-strategy is judged to be most effective at countering negative gender stereotypes. Our results indicate that 1) ChatGPT can be used to generate effective counter-stereotypes, 2) there are differences in terms of participant ratings depending both on whether the stereotype targets men or women and whether the participants themselves identify as male or female, and 3) the most promising strategies for future work are presenting counter-facts, or stating that all people can have the stereotypical trait regardless of their gender. The annotated dataset is publicly available.\footnote{\url{https://svkir.com/projects/gender-counter-stereotypes.html}} 

\vspace{5pt}

Our main contributions are as follows: 

\begin{itemize}
    \item A dataset presenting the ratings of offensiveness, plausibility, and effectiveness from 75 study participants for 220 counter-stereotypes. The counter-stereotypes are generated automatically according to 11 counter-strategies for 10 negative common stereotypes about men and 10 negative common stereotypes about women.
\item An annotation study indicating the most (and least) potentially effective counter-strategies to challenge gender-based stereotyping online.
\item Fine-grained analysis showing that while there are differences in the ratings depending on whether the participants identify as male or female, the more salient differences correspond to whether the stereotype targets men or women.
\end{itemize}

\section{Related Work}

This work draws on research from the social sciences as well as NLP and computer science. The processes by which stereotypes are formed, spread, and potentially disrupted have been studied extensively in social psychology, and we will provide only a brief overview of some key findings here. Researchers in NLP have recently begun translating these findings into new methods for countering hate speech, microaggressions, and stereotypes in online discourse, and this work will be discussed as well. 

\subsection{The Psychology of Stereotypes}

Stereotypes arise from cognitive processes that developed to help humans immediately categorize unknown strangers and determine their potential threat level (``friend'' or ``foe'') \cite{fiske2018model}. However, these cognitive shortcuts have limited utility in most modern social contexts and should be refined or discarded when additional information is available to make more accurate judgements about an individual. Nonetheless, gender stereotypes, whether consciously or subconsciously held, persist across different cultures and contexts.

Numerous studies evaluate different methods for reducing the effect of stereotypes.  Studies of racial bias have reported the effectiveness of strategies such as exposure to anti-stereotype exemplars (examples of people who disconfirm the stereotype in question) \cite{dasgupta2001malleability}, exercises that involve thinking from the target group’s perspective \cite{todd2011perspective}, and setting explicit goals for cooperation and equality \cite{blincoe2009prejudice, wyer2010salient}. 

\citet{palffy2023countering} conducted a field experiment to examine the effectiveness of counter-stereotypical framing and role models for adolescents choosing future occupations. They found that the intervention successfully increased the number of women who applied for typically male jobs in STEM fields. However, it did not increase the number of men who applied to typically female jobs in health and care-taking occupations. 

\citet{foster2022speaking} examined the question of how to reduce gender stereotypes in children, focusing on the \textit{essentialist} nature of such beliefs: the assumption that all members of a group are fundamentally the same due to some underlying essential nature. They observe that the statement ``Girls can be good at math too'' is a common response to the stereotypical statement of ``Boys are good at math.'' However, it only challenges the content of the specific claim about math skills while not addressing (and possibly even reinforcing) the essentialist belief that gender is a meaningful way to categorize people. They suggest instead the strategies of narrowing the scope of the statement (``Well, \textit{John} is good at math'') or broadening the scope of the statement (``Well, \textit{lots of kids} are good at math''), and show that these strategies are more effective at reducing prescriptivist beliefs about gender in 4-year-old children.

\subsection{Countering Stereotypes with NLP}

The NLP community began exploring the automatic generation of \textit{counter-speech} (statements challenging hate speech) with the work of \citet{qian-etal-2019-benchmark}, \citet{mathew2019thou}, and \citet{chung-etal-2019-conan} and followed by studies by \citet{tekiroglu-etal-2020-generating} and \citet{zhu-bhat-2021-generate}, among others. While still an active field of research, a new branch also recognizes the need to counter less extreme forms of abuse, such as stereotyping and microaggressions. Critically, while toxic content classified as ``hate speech'' can usually be removed from an online platform according to the terms of service, stereotyping and microaggressions are more likely to remain visible on the platform and thus need to be handled differently in order to mitigate their potential harms. Additionally, in many cases, the writers of such content have no intent to offend anyone, and therefore, there is also a component of education and empathy that can be useful in such cases.  

\citet{ashida-komachi-2022-towards} automatically generated counter-speech as well as `micro-interventions', a term which specifically refers to a statement countering a microaggression. They compared few-shot versus zero-shot approaches using GPT-2, GPT-3, and GPT-neo, and found that GPT-3 produced the least offensive and most informative responses, although they caution that fact-checking is necessary to avoid hallucinations or misinformation.

\citet{allaway2022towards} examined five strategies for countering essentialist claims in generic statements about groups. In line with \cite{foster2022speaking} above, they found that broadening statements, which remind the reader that these characteristics are not unique to one particular group, were rated quite highly. In contrast, annotators did not generally like direct counter-evidence, partly due to a high number of incorrect or subjective examples in the automatically generated text.

\citet{fraser-etal-2023-makes} surveyed the literature and identified 11 strategies for countering stereotypes that could potentially be implemented using generative language models. They used ChatGPT to automatically generate counter-stereotypes for 18 stereotypes common in North America, spanning the dimensions of negative--positive, descriptive--prescriptive, and ``more'' accurate--``less'' accurate. In a small ($n=4$, all female) annotation study, they determined that the strategies of denouncing stereotypes, warning of consequences, and using an empathetic tone were the most promising strategies overall. 
Another important finding of that study was that different strategies may be needed for different types of stereotypes (e.g., prescriptive versus descriptive or positive versus negative). We make use of \citeauthor{fraser-etal-2023-makes}'s list of counter-strategies, but focus specifically on negative, descriptive, gender-based stereotypes, one of the most prevalent types on social media. We conduct a much larger and more representative annotation study ($n=75$, 50\% male--female split), which allows us to also examine whether the perception of counter-stereotype strategies varies across the two genders.

\begin{table*}[!t]
\begin{center}
\begin{tabular}{ll}

      \hline
      \textbf{Stereotypes about women} & \textbf{Stereotypes about men}\\
      \hline
Women are bad at science. 	& Men are aggressive.\\
Women are irrational.	& Men only care about sex.\\
Women aren't strong leaders.	& Men are insensitive.\\
Women do not drive well.	& Men always feel the need to be dominant. \\
Women are not very assertive. 	& Men care more about their work than their families. \\
Women are submissive. 	& Men hate to show emotion.\\
Women are hysterical. 	& Men are egoistic. \\
Women are emotionally unstable. 	& Men lack empathy for others. \\
Women always cry. 	& Men don't care for others. \\
Women are dependent. 	& Men are uncomfortable with intimacy. \\
      \hline

\end{tabular}
\caption{Binary gender stereotypes used in this study. }
\label{tab:stereotypes}
 \end{center}
\end{table*}

\section{Study Design}

We describe next the study design: stereotype selection, counter-stereotype generation using automatic means (ChatGPT), and a questionnaire to gather participants' perceptions of counter-stereotype effectiveness.\footnote{This study was approved by the Ethics Boards at the authors' institutions.}

\subsection{Gender Stereotypes}

In this study, we focus on negative, descriptive, binary gender stereotypes (i.e., stereotypes portraying either men or women in a negative way), common in North America. 
We compiled a list of ten well-known stereotypes for each gender.  To compile the list of stereotypes, we selected the gender stereotypes that are well-known and grounded in existing literature, focusing on categories such as sociability/interpersonal connection and competence/professional achievement. We ensured that the stereotypes were unambiguously negative. Also, we aimed to make the sentences linguistically diverse. Then, the authors of the paper ranked those stereotypical views according to their prevalence in North American society. After aggregating the rankings, we chose the top 10 stereotypical views per gender to use in our experiments.
Table~\ref{tab:stereotypes} shows the stereotypes selected for the study.

\subsection{Generating Counter-Stereotypes}
\label{sec:generation}

Following \citet{fraser-etal-2023-makes}, we evaluate eleven counter-stereotype strategies:

\begin{enumerate}
    \item \textbf{Broadening exceptions:} Stating that the stereotypical trait is not unique to the target group by providing examples of other socio-demographic groups that share the trait.
    
    \item \textbf{Broadening universals:} Stating that the stereotypical trait is not unique to the target group and that all people, regardless of group membership, can have the trait.

    \item \textbf{Warning of consequences:} Pointing out possible negative outcomes of perpetuating the stereotype for the speaker, the target group, or the society.  
    
    \item \textbf{Counter-examples}: Providing examples of individuals or subgroups from the target group who do not have the stereotypical trait.
    
    \item \textbf{Counter-facts:} Providing facts that contradict the stereotype.
    
    \item \textbf{Critical questions:} Asking questions to motivate the speaker to review and possibly reconsider their beliefs.  
    
    \item \textbf{Denouncement:} Pointing out that the statement is a stereotype and perpetuating stereotypes is wrong.  
      
    \item \textbf{Empathy for the speaker:} Expressing empathy with the speaker's feelings and thoughts. 
    
    \item \textbf{Humour:} Challenge the stereotype using humour.  
    
    \item \textbf{Perspective-taking:} Asking the speaker to consider the stereotype from the target group's perspective. 
    
    \item \textbf{Emphasizing positive qualities:} Highlighting the positive characteristics of the target group. 
\end{enumerate}    

For each strategy and each stereotype, we prompted ChatGPT\footnote{\url{https://platform.openai.com/docs/models/gpt-3-5}} to generate one sentence in a social-media style. 
We used the prompts provided by \citet{fraser-etal-2023-makes}. 
In total, 220 counter-statements were generated. 

Next, we manually checked each counter-statement and excluded 31 statements that were not countering a given stereotype or that were generated using a strategy other than requested. 
For example, given the stereotype ``\textit{Men are egoistic''},
the statement ``\textit{Research shows that women and men have equal levels of self-esteem, and that men who expressed vulnerability were actually more well-liked than those who did not. \#byeegoisticstereotype}'' was rejected since it did not counter the corresponding stereotype. Likewise, ``\textit{Men do experience emotions, but societal expectations often discourage them from showing vulnerability and expressing themselves. \#empathyforall}'' was rejected as not showing empathy for the speaker when it was the requested strategy. 

Table~\ref{tab:wrong_strategy} shows the breakdown of the rejected counter-statements by strategy. 
Overall, ChatGPT was generally able to successfully generate counter-stereotypes for all strategies, except `broadening exceptions'. 
For most of the stereotypes, in place of `broadening exceptions', ChatGPT used a related strategy of  `broadening universals'. We speculate that ChatGPT chose this strategy 
to avoid making negative statements about particular social groups. 
Since 80\% of the counter-statements for `broadening exceptions' were generated using incorrect strategy, we decided to exclude this strategy from further study, leading to the exclusion of 35 counter-stereotypes in total. 
The remaining 185 counter-stereotypes for 10 strategies were presented to the participants in the survey.  




\begin{table}[!t]
\centering
\begin{tabularx}{0.8\columnwidth}{Xr}
      \toprule
      \textbf{Strategy} & \textbf{\# rejected} \\
      \midrule
      Broadening exceptions & 16 \\
      Broadening universals & 2 \\
      Consequences & 0 \\
      Counter-examples & 2 \\
      Counter-facts & 3 \\
      Critical questions & 0 \\
      Denouncement & 1 \\
      Empathy with speaker & 4 \\
      Humour & 1 \\
      Perspective-taking & 2 \\
      Positive qualities & 0 \\
      \midrule
      \textbf{Total} & \textbf{31} \\
      \bottomrule
\end{tabularx}
\caption{The number of counter-statements rejected after manual assessment. 
}
\label{tab:wrong_strategy}
\end{table}

\subsection{Questionnaire}
\label{subsec:questionnaire}

After providing informed consent, the participants were first presented with general instructions about the task. They were told that they would be shown gender stereotypes accompanied by the corresponding counter-stereotypes. Counter-stereotype was defined as follows: \textit{``A counter-stereotype challenges a gender stereotype. E.g., a counter-stereotype could present factual arguments against the gender stereotype, provide counter-examples or ask the speaker how they would feel if they were part of the target group. A counter-stereotype is \textbf{not} just the opposite of a gender stereotype.
''} Also, one example pair of stereotype--counter-stereotype was shown:

 \fbox{%
\parbox{0.43\textwidth}{%
Stereotype: \textit{Women are natural caretakers.}\vspace{2mm}\\
Counter-stereotype: \textit{I understand why some people may believe that women are natural caretakers, but gender does not determine one's ability or inclination to provide care. \#EndGenderStereotypes}
}%
}

\vspace{2mm}

\begin{figure}[t!]
    \centering
    \adjustbox{frame, center}{\includegraphics[width=\linewidth]{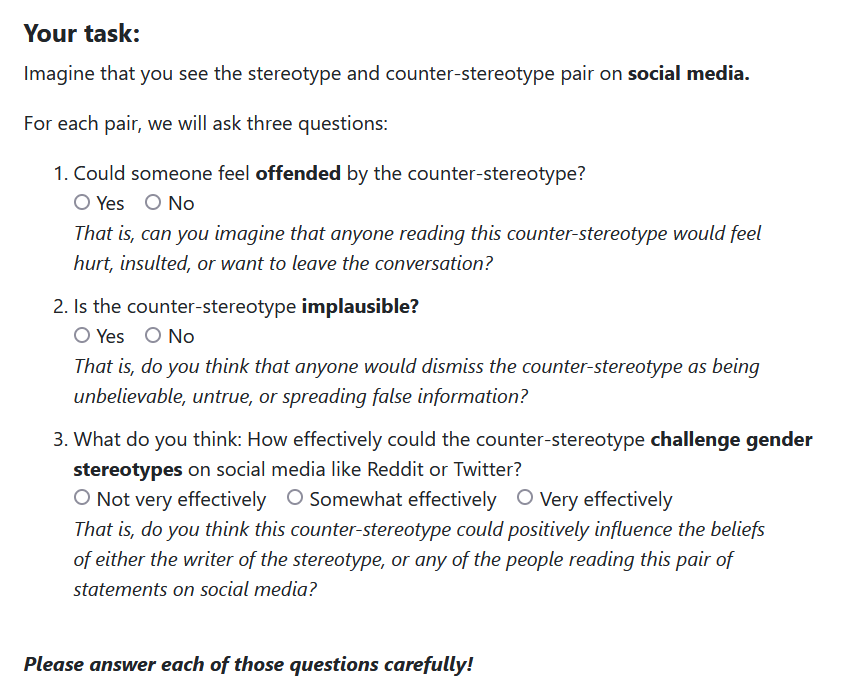}}
    \caption{The description of the task presented to study participants. }
    \label{fig:description}
\end{figure}

Then, the annotation task was presented and explained as shown in Figure~\ref{fig:description}. 
The task included questions to evaluate the \textbf{offensiveness}, \textbf{plausibility}, and potential \textbf{effectiveness} of each pair of stereotype--counter-stereotype. 
The questions related to offensiveness and plausibility required a binary answer, `yes/no'. 
The third question had three options: 
(1) very effectively (assigned a score of 1), (2) somewhat effectively (score of 0), (3) not very effectively  (score of -1).

To ensure the quality of responses, we employed three strategies: attention check questions, monetary bonus incentives, and monitoring the amount of time spent by a participant on the task. 

\subsection{Participants}

We recruited a total of 75 participants using \textit{Prolific}.\footnote{\url{https://www.prolific.co}} 
Since our study focuses on stereotypes prevalent in North America, we recruited participants solely from the U.S. and requested fluency in English.  
Out of 75 participants, 37 were male and 38 were female.\footnote{We used the gender specification self-reported by participants when signing up to Prolific; only binary options were available.} The mean (median) age was 40.41 (38) years. 
Each participant was asked to assess 30 stereotype--counter-stereotype pairs, which, on average, took around 15 minutes to complete. The participants were paid 3.00 USD (around \$12 per hour), which corresponds to the average reimbursement participants receive on Prolific. 
All participants passed the attention check, and each stereotype--counter-stereotype pair was rated on average by 12 participants (6 male and 6 female).

\section{Results}
In the following sections, we present and discuss the results of human rating for ten counter-strategies. (Recall that the strategy `broadening exceptions' was excluded due to ChatGPT's inability to consistently generate statements using this strategy (Sec.~\ref{sec:generation})).

\subsection{Offensiveness and Implausibility of Counter-Stereotypes}

Many counter-stereotypes were rated as potentially offensive by at least some participants.  On average, a counter-stereotype was perceived as potentially offensive by 35.8\% of participants who rated it, and 35 (out of 185) counter-stereotypes were perceived as potentially offensive by more than half of the raters. Counter-statements generated for the humour strategy were rated as offensive most frequently. 

These numbers are surprisingly high, especially taking into consideration the amount of effort put into guard-railing ChatGPT from generating offensive outputs. To clarify, none of the generated statements were explicitly offensive,\footnote{We consider a text explicitly offensive if it includes direct and unambiguous words or expressions, such as overtly derogatory language, intended to insult, degrade, or belittle someone.} yet participants indicated that some statements could potentially offend or upset certain social groups or users (e.g., \textit{'Just watched my husband try to fold a fitted sheet - dominance is not in his DNA. \#NoDominateGene \#HumorWins}, which was generated for the strategy `humour'). 

In fact, we note that there is a delicate boundary between what is considered offensive and non-offensive in this context. Even minor nuances can significantly influence how participants perceive a counter-statement. For example, the counter-stereotype \textit{``Women are strong, resilient, intelligent, nurturing, ambitious, and capable leaders who can handle their emotions without shedding a tear. \#WomenEmpowerment''} was perceived as offensive by the majority of the participants, but the sentence \textit{``Women are intelligent, intuitive, and capable decision-makers who excel in both emotional intelligence and logical reasoning. \#WomenAreNotIrrational''} was rated as non-offensive by all participants. Though seemingly positive, the former example promotes a narrow understanding of strength and leadership, implying that showing emotion or shedding a tear is a sign of weakness or incapability. In contrast, the second sentence is more inclusive and acknowledges a broad range of capabilities in women without falling into the prescribing of certain emotional responses. 
This observation highlights the complexity and the importance of validating the non-offensiveness of counter-stereotypes to ensure that counter-speech results in positive exchanges and does not escalate conflicts further. 

A counter-stereotype was perceived as implausible on average by 16.7\% of participants who rated it; six statements were rated as implausible by more than 50\% of participants.
We found a correlation between the offensiveness and implausibility ratings: counter-stereotypes that participants perceived as implausible were also often rated as potentially offensive (Pearson $\rho$ of 0.32). Counter-stereotypes perceived as offensive or implausible were also often rated as ineffective (Pearson $\rho$ of -0.21 and -0.29, respectively).

\subsection{Potential Effectiveness of Counter-Stereotypes}

We measure the potential effectiveness of counter-stereotypes by averaging the scores (1, 0, or -1) obtained by converting participants' answers to numerical values as described in Section \ref{subsec:questionnaire}.  



\begin{figure}[t]
\begin{center}
\includegraphics[width=\linewidth, trim = 2cm 0 0 0, clip]{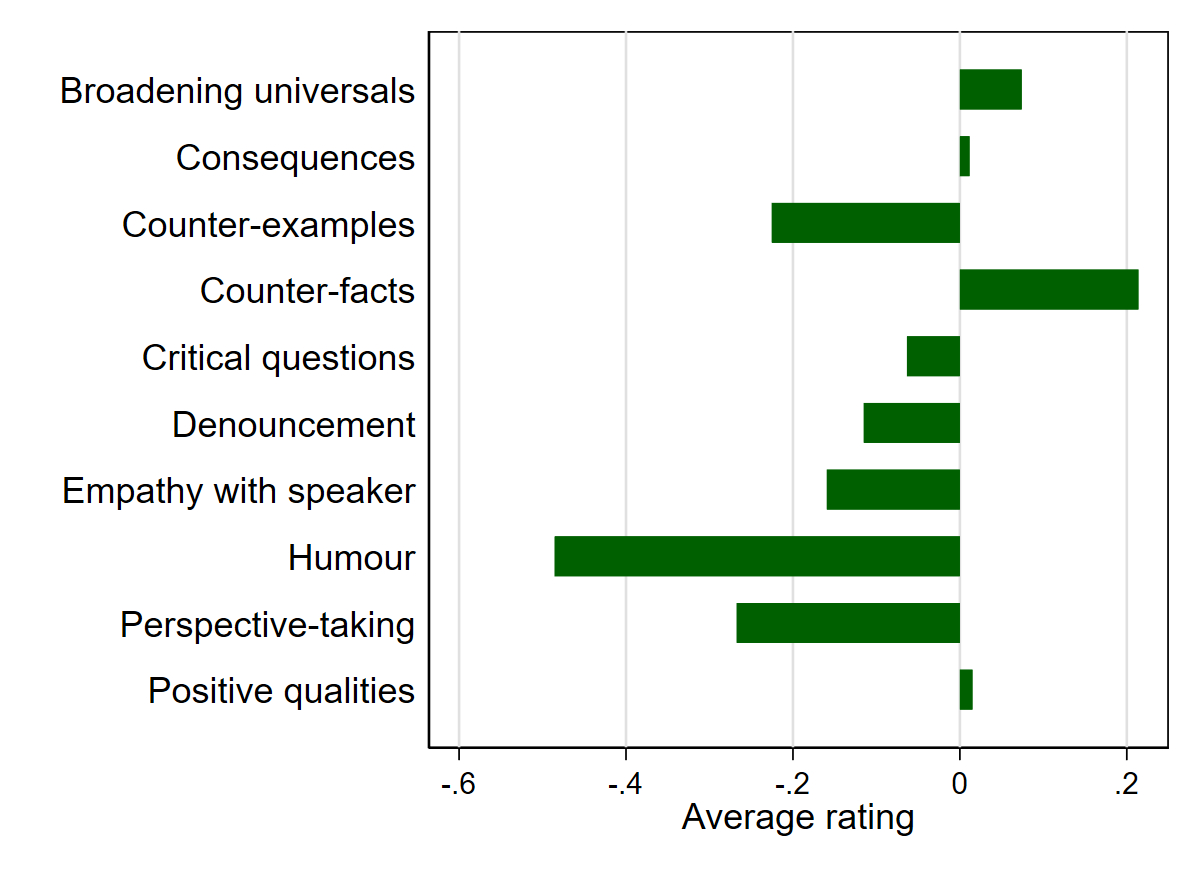} 

\caption{Average ratings of potential effectiveness for the ten counter-strategies.}
\label{fig:ratings-overall}
\end{center}
\end{figure}

\begin{figure*}[t!]
    \centering
    \begin{minipage}[c]{0.65\linewidth}        \includegraphics[width=\linewidth, trim={6cm 8.2cm 3cm 1cm}, clip]{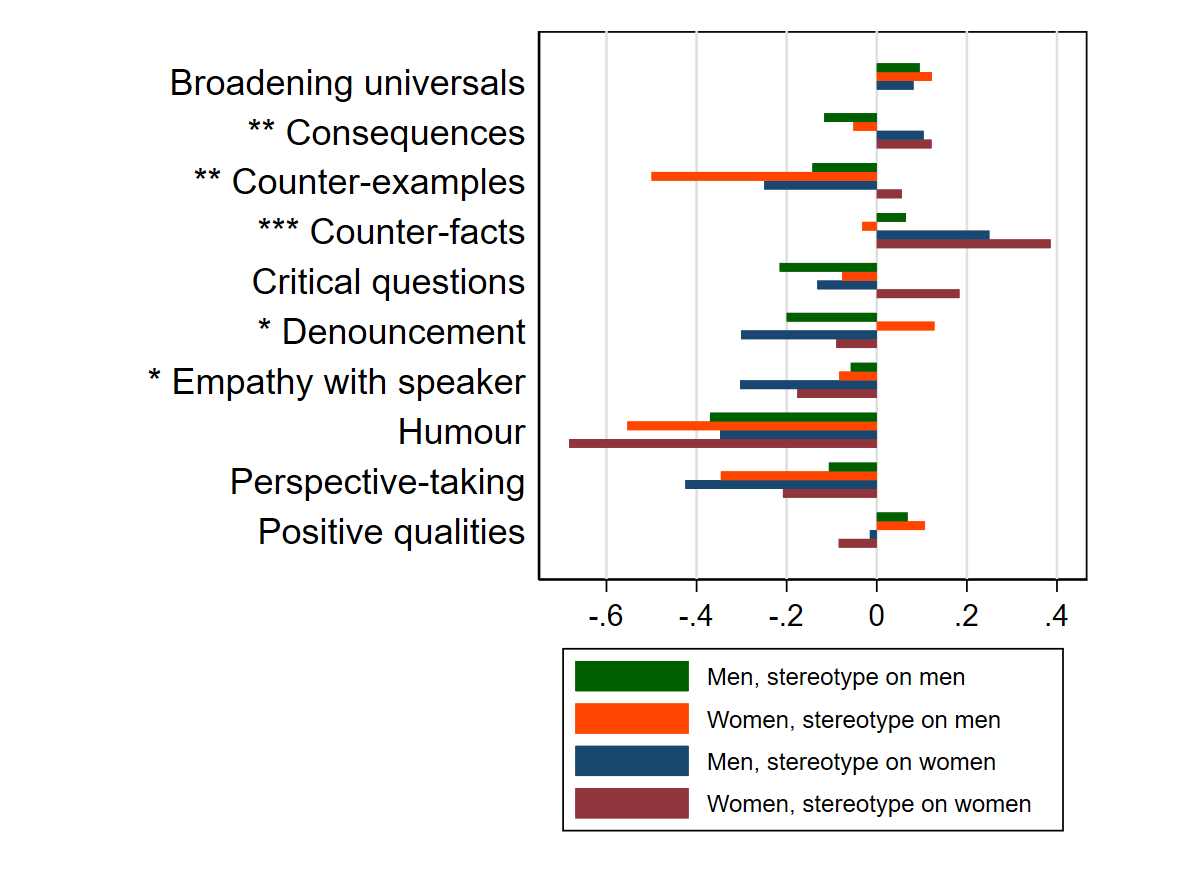}
    \end{minipage}%
    \begin{minipage}[c]{0.3\linewidth}        
        \centering        \includegraphics[width=\linewidth, trim={18cm 1cm 4.8cm 22.5cm}, clip]{figures/Graph_Subgroup.png}
    \end{minipage}
    \caption{Average ratings of potential effectiveness broken down by the participants' gender and the group the stereotype is about (men/women). For example, ``Men, stereotypes on men'' refers to male participants' ratings for counter-stereotypes about men. Statistically significant differences between ratings of stereotypes about men and women are marked as *~($p<0.1$), **~($p<0.05$), ***~($p<0.01$).}
    \label{fig:whole_figure}
\end{figure*}

\noindent{\textbf{Most and least effective strategies: }} Figure~\ref{fig:ratings-overall} shows the average ratings provided by the participants for each counter-stereotype strategy. Overall, two strategies, `counter-facts' and `broadening universals', received the highest positive ratings. 
The strategies of `emphasizing positive qualities' and `warning of consequences' received slightly positive average ratings, while all the other strategies were ranked negatively, on average. 
`Humour' stands out as the most potentially ineffective counter-stereotype strategy when automatically implemented by prompting ChatGPT. 

\vspace{5pt}

\noindent{\textbf{Fine-grained results across subgroups: }} To get further insights, we split the ratings by the stereotyped group (stereotypes about women vs. stereotypes about men) and by participant-reported gender (male vs. female). 
Figure~\ref{fig:whole_figure} shows the average ratings per subgroup and Table ~\ref{tab:all-results} shows the differences in average ratings by stereotype target and by participants' gender. 

Overall, we observe greater differences in the ratings for 
stereotypes about men versus stereotypes about women (mean absolute difference of 0.15) than between the ratings of male and female participants (mean absolute difference of 0.10). 
For example, although `counter-facts' received mostly positive ratings for both subgroups of stereotypes, statements countering 
stereotypes about women were rated substantially higher than statements countering 
stereotypes about men, regardless of the gender of the participant. 
Interestingly, female participants perceived counter-facts opposing 
stereotypes about men (e.g., \textit{``According to a study by the American Psychological Association, men reported higher levels of intimacy overall, including emotional and physical intimacy, than women did. \#menareintimate \#breakthestereotype''}) as offensive and/or implausible at substantially higher rates than counter-facts opposing 
stereotypes about women. This was also at higher rates than male participants perceived counter-facts opposing stereotypes about both groups. 
Also, `emphasizing positive qualities' was mostly perceived as effective for countering 
stereotypes about men, while `warning of consequences' received higher ratings for 
stereotypes about women.

The strategies showing the greatest difference in ratings between male and female participants were `critical questions' and `denouncement', both of which were preferred more by female participants, and `humour', which was disliked by all participants, but more so by females. 
`Perspective-taking' and `counter-examples' were rated higher by female participants for stereotypes about women and by male participants for stereotypes about men.

\setlength{\tabcolsep}{2pt}
\begin{table}
\centering
\begin{tabular}{lrlrl}
\toprule
Strategy & \multicolumn{2}{l}{Diff. by} & \multicolumn{2}{l}{Diff. by partic.} \\
& \multicolumn{2}{l}{target} & \multicolumn{2}{l}{gender}\\
\midrule
Broadening universals & - 0.07 & &- 0.03 &\\
Consequences & 0.20 &** & 0.05& \\
Counter-examples & 0.26&** & -0.04&\\
Counter-facts & 0.30&*** & 0.05&\\
Critical questions & 0.17 & & 0.22& **\\
Denouncement & - 0.14&* & 0.26&*** \\
Empathy with speaker& - 0.17&* & 0.05 &\\
Humour & -0.04 && - 0.24&**\\
Perspective-taking & -0.07& & -0.03 &\\
Positive qualities & - 0.14 && - 0.03&\\
\midrule
Avg. absolute diff. & 0.15 && 0.10 &\\
\bottomrule
\end{tabular}
\caption{The differences in avg. ratings by stereotype target (avg. rating of counter-stereotypes about women $-$ avg. rating of counter-stereotypes about men)  and by participants' gender (avg. rating by women $-$ avg. rating by men), per strategy. Statistically significant differences are marked as *~($p<0.1$), **~($p<0.05$), ***~($p<0.01$). }
\label{tab:all-results}
\end{table}

\setlength{\tabcolsep}{6pt}

\section{Discussion}

Our research indicates a clear distinction in the potential effectiveness of various strategies when generated automatically. Certain strategies have a more universal appeal, 
while others are perceived as having limited or even negative effect. 

`Counter-facts' and `broadening universals' were the two strategies that exhibited the most positive outcomes across different scenarios and audiences. `Broadening universals' involves presenting a wider, more inclusive understanding of a particular trait or behaviour, challenging the limited scope of stereotypes. Similarly, counter-facts provide a logical and evidence-based method of disputing any unfounded claims. These approaches immediately offer an alternative viewpoint, urging individuals to reconsider their biases. However, we observed that female participants sometimes found the counter-facts ineffective when used to counter stereotypes about men. 
This suggests that a one-size-fits-all approach may be insufficient and audience-specific nuances, where applicable, may enhance the effectiveness of a strategy. 

On the other hand, strategies such as `humour', `perspective-taking', `counter-examples', and `empathy with the speaker' 
have been rated as ineffective. Among these, the use of `humour' in counter-statements was particularly problematic. While humour can be a powerful tool for challenging societal norms, it also risks being offensive or misinterpreted, especially in sensitive areas like gender stereotypes. Also, producing high-quality humour is a difficult technical task for NLP models; even state-of-the-art generative models have yet to master this task.  In our findings, automatically generated humorous counter-statements were interpreted as offensive and implausible by a significant portion of the audience. This highlights the precarious nature of using humour in automatic interventions; it can inadvertently reinforce the very stereotypes it aims to challenge.

Overall, we observe that, on average 35.8\% of participants rated AI-generated counter-stereotypes as potentially offensive, emphasizing the complexities of automating responses in delicate subject areas like gender stereotypes. Besides humorous statements,  counter-strategies that challenge a stereotype about one binary gender by directly attributing the negative trait to the opposite gender (e.g., by providing statistics on the other gender) may be seen as unnecessarily confrontational or belittling. 

Furthermore, strategies like `perspective-taking', which involves encouraging individuals to view a situation from another person's standpoint, and `empathy', which focuses on fostering a deeper emotional understanding, fail to resonate with the audience in case of online gender stereotyping. While these strategies are valuable in interpersonal interactions, they might be interpreted as irrelevant, insincere or patronizing when automated and used in online communications.

Another finding in our data is the noticeable difference in ratings when comparing stereotypes about women versus stereotypes about men. We observe that the stereotype's inherent nature and societal connotations play a larger role than the gender of the individual evaluating it. 
Due to a long history of the oppression of women and contemporary movements for reclaiming women's rights, societal discourse, both offline and online, includes extensive discussions on the issue. This has allowed models like ChatGPT to be adept at generating `counter-facts' and `warnings of consequences' that directly address these stereotypes. In other words, the established nature of stereotypes about women, paired with the rich database of discussions and rebuttals around them, appears to enable AI models to craft plausible and effective responses to counter such stereotypes.

In contrast, stereotypes that target men present a challenge for automatic countering. The counter-facts generated in response to stereotypes about men were often deemed less effective by participants. For example, in response to the stereotype \textit{``Men don't care for others''}, ChatGPT generated the counter-fact, \textit{``According to a study by the DoSomething.org, men are actually more likely than women to donate to charity and volunteer their time for causes. \#MenCareToo''}, which may be perceived as implausible or offensive by women. 
However, in response to the stereotype \textit{``Women are irrational''}, it generates the counter-fact \textit{``Studies show that men and women make decisions with similar levels of emotional involvement and rational thinking, debunking the stereotype that women are irrational. \#GenderBias \#FactsOverStereotypes''}, which has a neutral tone about both genders. Also, our qualitative assessment shows that counter-stereotypes generated for men are less specific than those generated for women. For example, for the strategy of mentioning the consequences of stereotypes, to counter the stereotype \textit{``Men are aggressive.''} ChatGPT generates \textit{``Spreading the stereotype that men are aggressive can lead to harmful generalizations and discrimination, let's break the cycle. \#StopStereotyping.''}, but for stereotype \textit{``Women are irrational.''} it generates \textit{``Spreading the stereotype that women are irrational can lead to women being underestimated and undervalued, ultimately hindering progress and equality.''}. While the first statement broadly highlights the consequences of \textit{all} stereotypes, 
it does not specify who exactly would be harmed, or how. On the other hand, the second statement explicitly mentions that women will be the ones ``underestimated'' and ``undervalued'' due to the stereotype and specifies its larger societal impact, suggesting that such stereotypes can ``hinder overall progress and equality''. This might stem from the fact that discussions on stereotypes about men, while present, are not as prevalent or as deeply ingrained as their counterparts, stereotypes about women. Thus, traditional strategies, like counter-facts, were not ranked as effective. However, our research found that other approaches, like mentioning the positive qualities inherent in men and emphasizing the fact that traits like aggression and insensitivity can be found across all genders, were deemed more effective.




\section{Data Usability}

We make our annotated dataset publically available to foster future research. The dataset is in English and was developed to understand the perceptions of individuals regarding counter-statements to negative, descriptive gender stereotypes about men and women. The primary goal was to evaluate the offensiveness, plausibility, and potential effectiveness of counter-statements when presented in response to stereotyping in online platforms.



We anticipate several potential uses for this dataset. First, researchers can use this data to understand common gender stereotypes and the societal perception of counter-stereotypes. Second, this data might be used for NLP tasks such as training and/or evaluating models to identify, respond to, or counter stereotypes in digital content. Third, the insights learned from this data might be used by NGOs or community groups to craft more effective stereotype-countering campaigns.

The dataset comes with limitations that need to be understood and mitigated before considering the above use cases. The data includes only negative, descriptive, binary gender stereotypes. Therefore, it does not encompass all facets of stereotypes and is not inclusive of all gender identities. 
By nature, stereotypes can be sensitive and potentially offensive; care should be taken in their use and interpretation to avoid perpetuating harmful beliefs or norms. Further, the collected ratings capture perceptions towards stereotypes and their counter-stereotypes and do not assert the truth or validity of these statements. Previous research documented many cases of generative AI models hallucinating or fabricating information \cite{ye2023cognitive}, so even when deemed plausible, fact-checking of generated texts is essential to prevent misinformation. Finally, the ratings provided by the participants are subjective and reflect the participants' opinions, which might be affected by their personal experiences and beliefs. Further research is required to assess the actual effectiveness of these strategies in changing online users' stereotypical beliefs.


\section{Conclusion}
We conducted a large-scale online study on the potential effectiveness of automatically generated statements to counter common negative gender stereotypes. We found that while some strategies offer a promising avenue to counter stereotypes, others require careful consideration to ensure they don't have an adverse effect. 
Special attention needs to be paid to the delicate balance between challenging users' views and making them feel provoked or offended. 
Confronting gender stereotypes requires a nuanced and tailored approach, taking into account 
stereotypes' historical, cultural, and societal context. 
The effectiveness of a counter-strategy may vary according to the social group being stereotyped, the expected audience, and other contextual features. 

\section{Ethics Statement}

While our study represents a promising step in counteracting gender stereotypes online, it comes with limitations and ethical considerations that require ongoing attention. Emphasizing transparency, continuous evaluation, and social context will be essential as we navigate this intersection of technology and societal constructs.

We release a dataset that can be used for further research or social applications on countering stereotypes. Before using this dataset, users should familiarize themselves with the context and limitations of the dataset. The dataset should be handled and interpreted responsibly, considering the potential ethical implications \cite{kirk2022handling}. We also strongly recommend cross-referencing with other relevant literature or datasets for a more holistic view. It is important to note that the participants’ biases might influence their perceptions and judgements, potentially reflecting in the individual ratings. In our study, we rely on the fact that these biases are likely to affect all the strategies uniformly and report the statistically significant differences between the counter-stereotype strategies. Therefore, while the relative comparisons between strategies remain valid and meaningful, individual ratings should be used with care.

One limitation of this research is its focus on binary gender stereotypes, including only male and female identities. We are fully cognizant of the prevalence and significance of stereotypes about non-binary individuals in contemporary online discourse and lived experiences. Our decision 
stems not from oversight but from an understanding of the intricate complexity of the issue, 
 which warrants an exhaustive and dedicated study separate from the constraints of the present research.

 If successful, automatic methods to counteract gender stereotypes can reshape online conversations, fostering more inclusive and equal digital environments. Such interventions can be instrumental in actively challenging entrenched biases, potentially influencing societal perceptions and behaviours. However, while AI models can mimic human language patterns, they may not always capture the subtleties and sensitivities needed when addressing deeply embedded societal constructs, so human oversight might be necessary, especially in critical applications. Furthermore, our sample's demographics might not be globally representative, limiting the generalizability of the findings.

As indicated, certain counter-strategies, especially when applied across genders, heightened perceptions of offensiveness and implausibility. These interventions could inadvertently perpetuate biases or spark unintentional controversies without careful calibration. It is essential to take caution, ensuring strategies do not polarize views further or cause distress to the audience. Also, even while attempting to challenge stereotypes, current AI models can inadvertently reproduce or reinforce societal biases present in the training data. Acknowledging this limitation and working continually to minimize such repercussions is crucial.

AI-based interventions need consistent evaluation and refinement. What works today might not be as effective tomorrow due to evolving societal contexts. Regular reassessment ensures interventions remain relevant and impactful. It is imperative that users are informed when AI-generated outputs are being employed to counteract stereotypes, supporting transparent online interactions.


\section{Bibliographical References}
\label{reference}

\bibliographystyle{lrec-coling2024-natbib}
\bibliography{references}


\end{document}